\documentclass[11pt]{article}

\usepackage[margin=1in]{geometry}
\usepackage{graphicx}
\usepackage{booktabs}
\usepackage{amsmath,amssymb}
\usepackage[authoryear,round]{natbib}
\setcitestyle{semicolon,aysep={ }}
\usepackage[font=small,labelfont=bf]{caption}
\usepackage{placeins}
\usepackage[hyphens]{url}
\usepackage[hidelinks]{hyperref}
\usepackage{microtype}

\title{\bfseries SurgGMF: Fully Causal Gaussian Motion Forecasting for Anticipatory Surgical Scene Rendering}

\author{
Jingqian Sun$^{1}$ \qquad Yichao Tang$^{1,2,3,*}$\\[0.5em]
\small $^{1}$Shanghai Research Institute for Intelligent Autonomous Systems, Tongji University, Shanghai, China\\
\small $^{2}$School of Mechanical Engineering, Tongji University, Shanghai, China\\
\small $^{3}$Shanghai Innovation Institute, Shanghai, China\\[0.25em]
\small $^{*}$Corresponding author: \texttt{tangyichao@tongji.edu.cn}\\
\small \texttt{jingqian\_sun@tongji.edu.cn, tangyichao@tongji.edu.cn}
}
\date{}

\begin{document}

\maketitle

\begin{abstract}

Dynamic surgical scene modeling is essential for robotic perception, simulation, and decision support. Although existing neural rendering methods enable efficient reconstruction and rendering of deformable surgical scenes, they remain primarily focused on observed-frame reconstruction rather than forecasting future scene states. To this end, we present SurgGMF, a fully causal Gaussian motion forecasting framework for anticipatory surgical scene rendering. Rather than predicting future RGB images directly, SurgGMF forecasts future Gaussian motion states represented by position, scale, and rotation residuals (X/S/R) from historical Gaussian motion fields. To prevent target leakage, we introduce a full-causal-last rendering protocol, where future Gaussian states are rendered without accessing target-frame Gaussian attributes while preserving causal appearance propagation. We evaluate SurgGMF on 12 EndoNeRF and StereoMIS video slices using neural temporal learners and classical dynamics baselines under a unified forecasting protocol. Learned Gaussian motion forecasting consistently outperforms classical dynamics baselines in render space, demonstrating gains beyond hand-crafted state extrapolation. Latency analysis further reveals an accuracy--efficiency trade-off: under the current implementations, TKAN achieves the highest accuracy, whereas GRU and LSTM provide more favorable module-level latency profiles.  These results establish SurgGMF as a reproducible framework for causal Gaussian motion forecasting and advance surgical Gaussian representations from retrospective reconstruction toward predictive scene modeling.

\end{abstract}

\section{Introduction}

Robot-assisted and image-guided surgical systems increasingly require predictive models of dynamic surgical scenes. In minimally invasive surgery, reconstruction remains challenging due to nonrigid deformation, partial visibility, weak texture, specular reflection, and tool occlusion \citep{ozyorukEndoSLAMDatasetUnsupervised2021,wangNeuralRenderingStereo2022,zhaEndoSurfNeuralSurface2023}. Since surgical reconstruction supports intraoperative navigation and robotic assistance \citep{wangNeuralRenderingStereo2022,zhaEndoSurfNeuralSurface2023}, representations that only reconstruct current or previously observed frames are insufficient for tasks requiring short-term anticipation. This motivates surgical scene representations that support both reconstruction and future state prediction.

Recent studies in world modeling and predictive representation learning demonstrate that future state rollout can be enabled by learning latent states and transition dynamics \citep{bardesRevisitingFeaturePrediction2024,bruceGenieGenerativeInteractive2024,hafnerMasteringDiverseControl2025}. However, direct pixel forecasting does not expose the structured 3D variables required for geometric reasoning, temporal tracking, and view-consistent rendering. This limitation is particularly relevant in surgical scenes, where deformable geometry is critical for consistent reconstruction and interaction. 3D Gaussian splatting (3DGS) provides a natural state representation for this objective by offering a renderable Gaussian-based state space \citep{kerbl3DGaussianSplatting2023}. Recent dynamic and surgical Gaussian methods extend 3DGS to deformable and endoscopic scenes \citep{luitenDynamic3DGaussians2024,yangDeformable3DGaussians2024,huangEndo4DGSEndoscopicMonocular2024,xieSurgicalGaussianDeformable3D2024,yangDeform3DGSFlexibleDeformation2024}. However, existing dynamic Gaussian approaches mainly focus on scene reconstruction, motion modeling, or general-purpose extrapolation, while causal forecasting of surgical Gaussian states under strict target-frame isolation remains underexplored.

We introduce SurgGMF, a fully causal Gaussian motion forecasting framework for anticipatory surgical scene rendering. SurgGMF forecasts future Gaussian motion states rather than future RGB images. Given historical Gaussian motion fields, it predicts future Gaussian position, scale, and rotation residuals (X/S/R) and renders future frames by updating Gaussian states. To avoid target leakage, we define a full-causal-last protocol in which predicted frames are rendered without access to target-frame Gaussian attributes; appearance and uncovered primitives are inherited only from the last causally available frame. The goal of SurgGMF is not to identify a single optimal temporal learner, but to establish and analyze a causal surgical Gaussian forecasting formulation. We therefore systematically evaluate neural temporal learners against controlled classical dynamics baselines, analyze Gaussian motion component contributions, and study accuracy--efficiency trade-offs. Our contributions are:

\begin{itemize}
\item We formulate causal Gaussian motion forecasting for anticipatory surgical rendering, predicting future X/S/R Gaussian residuals from historical motion fields without synthesizing future RGB directly.
\item We introduce a full-causal-last rendering protocol that prevents access to target-frame Gaussian attributes and enables reproducible evaluation of causal Gaussian forecasting.
\item We systematically evaluate neural temporal learners and classical dynamics baselines under a unified X/S/R forecast-and-render protocol.
\item We analyze Gaussian motion component contributions and accuracy--efficiency trade-offs, showing the effectiveness of joint X/S/R prediction and revealing different accuracy-latency characteristics among temporal learners under module-level evaluation.
\end{itemize}

\section{Related Work}

\paragraph{Predictive World Models and Structured Scene Forecasting.}

World models learn state representations and transition dynamics for future rollout, imagination, and planning \citep{hafnerMasteringDiverseControl2025}. Existing approaches commonly predict latent features, visual observations, or action-conditioned trajectories \citep{bardesRevisitingFeaturePrediction2024,bruceGenieGenerativeInteractive2024,assranVJEPA2SelfSupervised2025,heVideoGenerationModels2026,mur-labadiaVJEPA21Unlocking2026}. Although effective for predictive representation learning, latent- or image-space states do not directly provide explicit geometry for spatial reasoning and view-consistent rendering. Recent autonomous-driving methods therefore introduce structured Gaussian states: GaussianWorld performs streaming 4D occupancy forecasting in a semantic Gaussian space, while GaussianAD predicts Gaussian flow for occupancy forecasting and planning \citep{zuoGaussianWorldGaussianWorld2025,zhengGaussianADGaussianCentricEndtoEnd2024}. These studies establish Gaussian primitives as predictive scene representations, but focus on semantic occupancy and road-scene dynamics rather than rendering continuously deforming surgical anatomy.

\paragraph{Surgical Reconstruction and Dynamic Gaussian Modeling.}

Neural scene representations have been widely studied for reconstructing deformable surgical environments. EndoNeRF and related neural radiance- or surface-field methods model nonrigid tissue deformation from monocular, stereo, or RGB-D endoscopic observations \citep{wangNeuralRenderingStereo2022,zhaEndoSurfNeuralSurface2023,yangEfficientDeformableTissue2024}. These approaches achieve high-quality reconstruction by jointly optimizing geometry, appearance, and deformation over an observed sequence, but their implicit representations do not naturally expose persistent primitive-level trajectories for downstream temporal forecasting.

3DGS provides an explicit alternative by representing geometry and appearance with anisotropic Gaussian primitives \citep{kerbl3DGaussianSplatting2023}. Dynamic extensions model temporal variation through persistent Gaussian trajectories, canonical-space deformation fields, or explicit spatiotemporal features \citep{yangRealtimePhotorealisticDynamic2023,wu4DGaussianSplatting2024,luitenDynamic3DGaussians2024,yangDeformable3DGaussians2024,liSpacetimeGaussianFeature2024}. Surgical Gaussian methods adapt these formulations to endoscopic scenes using tissue deformation models, depth supervision, tool-aware constraints, or online optimization \citep{liuFoundationModelGuidedGaussian2025,huangEndo4DGSEndoscopicMonocular2024,xieSurgicalGaussianDeformable3D2024,yangDeform3DGSFlexibleDeformation2024,hayozOnline3DReconstruction2024,paonimEndoPlanarDeformablePlanarBased2026,chenSurgicalGSDynamic3D2026}. Collectively, these approaches provide effective representations for reconstructing and tracking observed tissue motion, but generally do not evaluate future-state extrapolation under a fixed historical observation boundary.

\paragraph{Predictive Gaussian Dynamics and Temporal Modeling.}

Recent work directly predicts future states in Gaussian representations. GaussianPrediction transfers motion from a sparse graph of control keypoints to a dynamic Gaussian scene for motion extrapolation and future free-view synthesis \citep{zhaoGaussianPredictionDynamic3D2024a}. Graph-based Gaussian dynamics models similarly derive control particles from tracked Gaussians and learn action-conditioned object dynamics for manipulation and future rendering \citep{zhangDynamic3DGaussian2024}. Continuous-time approaches provide an alternative: ODE-GS evolves latent Gaussian trajectory representations with a neural ordinary differential equation, while ParticleGS models Gaussians as a particle system and learns differential dynamics for motion extrapolation \citep{wangODEGSLatentODEs2025,quanParticleGSLearningNeural2026}. These methods demonstrate that Gaussian states can support learned future extrapolation in general dynamic scenes.

SurgGMF differs primarily in its surgical forecasting formulation and causal evaluation boundary. It forecasts short-horizon position, scale, and rotation residuals from teacher-derived surgical Gaussian trajectories and evaluates all methods using the same full-causal-last renderer, without access to target-frame Gaussian attributes. Its focus is therefore controlled causal evaluation of Gaussian forecast-and-render pipelines for deformable surgical scenes rather than Gaussian extrapolation in general.

Recurrent models such as GRU and LSTM, attention-based Transformers, and TKAN provide complementary inductive biases for multi-step forecasting \citep{hochreiterLongShortTermMemory1997,choLearningPhraseRepresentations2014,vaswaniAttentionAllYou2017,genetTKANTemporalKolmogorovArnold2025}. we treat these architectures as interchangeable forecasting backbones and compare them with classical dynamics references in a common Gaussian residual space.

\section{Method}

\begin{figure}[t]
\centering
\includegraphics[width=\textwidth]{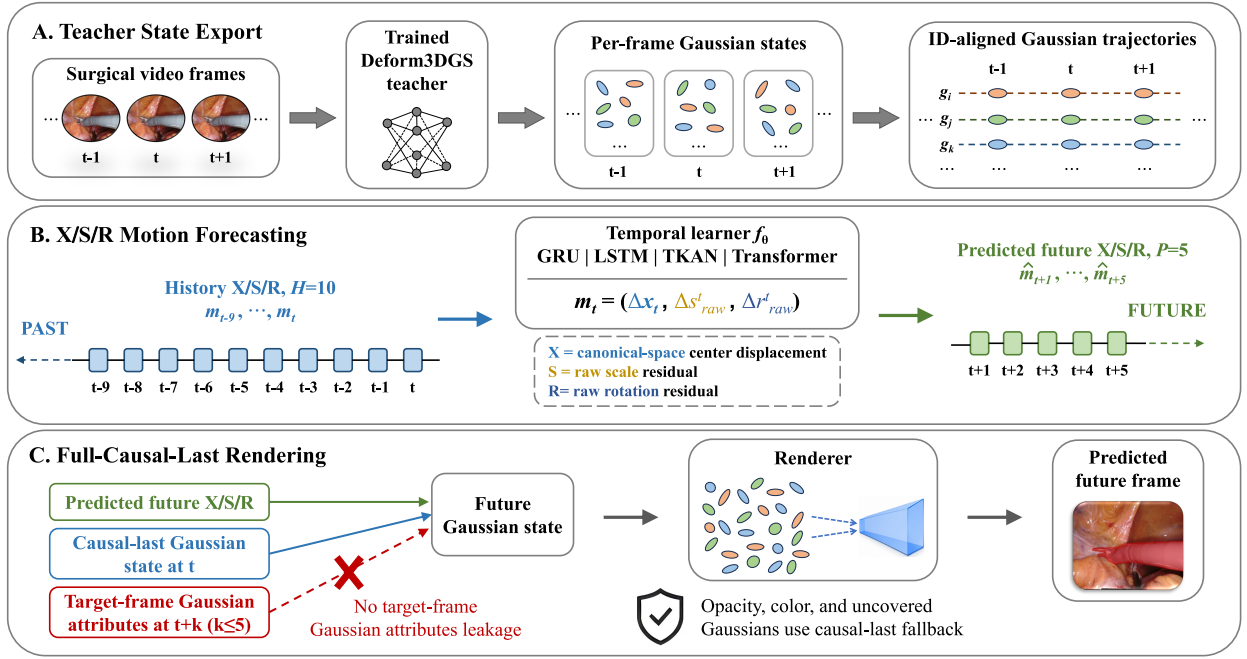}
\caption{Overview of SurgGMF. A trained Deform3DGS teacher exports temporally aligned Gaussian states. SurgGMF constructs X/S/R motion histories, forecasts future Gaussian motion states with a temporal learner or a dynamics baseline, converts predicted residuals back into future Gaussian states, and evaluates rendered future frames under the full-causal-last protocol.}
\label{fig:pipeline}
\end{figure}

In this study, we introduce SurgGMF, a fully causal Gaussian motion forecasting framework for anticipatory surgical scene rendering. The goal is to predict the future geometric evolution of a dynamic surgical scene representation and evaluate whether the predicted Gaussian states can support future-frame rendering without accessing target-frame Gaussian attributes. SurgGMF operates in a structured Gaussian state space derived from dynamic 3D Gaussian reconstruction, and separates geometric motion forecasting from appearance propagation under a strict causal rendering protocol. This separation allows future geometry to be evaluated independently of target-frame reconstruction. Figure~\ref{fig:pipeline} summarizes the overall pipeline.

\subsection{Causal Gaussian Motion Forecasting}

Let a dynamic surgical scene at time $t$ be represented by a set of 3D Gaussian primitives,
\[
\mathcal{G}^{t} = \{ g_i^{t} \}_{i=1}^{N_t},
\]
where each Gaussian primitive contains geometric and appearance-related attributes,
\[
g_i^{t} = (x_i^{t}, s_i^{t}, r_i^{t}, \alpha_i^{t}, c_i^{t}).
\]
Here, $x_i^{t}$ denotes the Gaussian center, $s_i^{t}$ denotes scale parameters, $r_i^{t}$ denotes rotation parameters, $\alpha_i^{t}$ denotes opacity, and $c_i^{t}$ denotes color/appearance features represented by spherical harmonic (SH) coefficients. In our implementation, these Gaussian states are teacher-derived states exported from a trained Deform3DGS model \citep{yangDeform3DGSFlexibleDeformation2024}. They provide a renderable and temporally aligned state space, but should not be interpreted as physical ground-truth tissue trajectories.

Temporal correspondence is established by a checkpoint-local Gaussian identifier. Since the visible Gaussian subset may vary across frames, SurgGMF constructs training samples only from Gaussian trajectories with complete, finite history and target windows. This yields temporally aligned Gaussian motion sequences without assuming that every frame contains the same primitive set.

Instead of directly predicting future RGB images, SurgGMF forecasts future Gaussian motion states. We define the X/S/R forecasting target as
\[
m_i^{t} =
(\Delta x_i^{t}, \Delta s_{i,\mathrm{raw}}^{t}, \Delta r_{i,\mathrm{raw}}^{t}),
\]
where X denotes the canonical-space center displacement, S denotes the raw scale residual, and R denotes the raw rotation residual. Specifically,
\[
\Delta x_i^{t} = x_i^{t} - x_i^{c},
\]
where $x_i^{c}$ is the canonical Gaussian center. The scale and rotation targets are raw residuals from the Deform3DGS deformation output before renderer-side activation. This definition treats Gaussian motion as the time-varying residual state of each primitive relative to its canonical configuration, rather than as a simple adjacent-frame displacement.

Given a history window of length $H$, SurgGMF predicts a future horizon of length $P$:
\[
\hat{m}_{i}^{t+1:t+P}
=
f_{\theta}
\left(
m_i^{t-H+1:t}
\right).
\]
The implementation augments this history with deterministic velocity and acceleration features, as detailed below. In the formal experiments, we use $H=10$ and $P=5$. As illustrated in Figure~\ref{fig:pipeline}B, a single forward pass jointly predicts all five future steps without autoregressive rollout or ground-truth future-state feedback.

\subsection{SurgGMF Framework}

SurgGMF consists of three stages: motion-sequence construction, temporal forecasting, and future Gaussian state construction.

\paragraph{Motion-sequence construction.}
Dynamic Gaussian states are converted into X/S/R motion sequences. For each temporally valid Gaussian trajectory, the historical input contains canonical-space center displacement, raw scale residuals, and raw rotation residuals. To provide local temporal cues, the input also includes first-order velocity and second-order acceleration derived from the historical X trajectory. In the X/S/R setting, each sample is represented as a sequence of length 10, and the per-Gaussian input contains 16 channels: three center-displacement channels, three velocity channels, three acceleration channels, three raw-scale residual channels, and four raw-rotation residual channels. The same temporally aligned history window is used by neural learners and by non-learning dynamics baselines, ensuring that all methods access the same causal information.

\paragraph{Temporal forecasting.}
A forecasting function $f_{\theta}$ predicts future X/S/R states. SurgGMF is model-agnostic with respect to the forecasting backbone. In this work, we instantiate it with recurrent, KAN-based, and attention-based learners: GRU, LSTM, TKAN, and Transformer models. All learners share the same input/output construction, X/S/R forecasting target, and downstream rendering protocol. We also evaluate non-learning dynamics baselines in the same X/S/R residual space. This controlled setting allows us to compare different temporal inductive biases while keeping the forecasting task and causal evaluation boundary fixed.

\paragraph{Future Gaussian state construction.}
Predicted X/S/R residuals are converted back into future Gaussian states. For a future step $t+k$, the predicted geometric attributes are obtained by adding the predicted residuals to the corresponding canonical raw Gaussian parameters:
\[
\hat{x}_i^{t+k}
=
x_i^{c}
+
\widehat{\Delta x}_i^{t+k},
\]
\[
\hat{s}_{i,\mathrm{raw}}^{t+k}
=
s_{i,\mathrm{raw}}^{c}
+
\widehat{\Delta s}_{i,\mathrm{raw}}^{t+k},
\]
\[
\hat{r}_{i,\mathrm{raw}}^{t+k}
=
r_{i,\mathrm{raw}}^{c}
+
\widehat{\Delta r}_{i,\mathrm{raw}}^{t+k}.
\]
The raw scale and rotation parameters are then passed to the renderer, where the standard Deform3DGS scale and rotation activations are applied. Thus, SurgGMF predicts raw residuals in the Deform3DGS parameter space, while validity constraints such as scale activation and rotation normalization remain handled by the original rendering pipeline.

\subsection{Full-Causal-Last Rendering Protocol}

A central issue in future Gaussian rendering is target-frame leakage. If target-frame Gaussian attributes are used to fill missing or non-forecasted states, the rendered future image may implicitly depend on information unavailable at prediction time. To avoid this, we define a full-causal-last rendering protocol.

For a target frame $T$ and horizon step $k$, the causal fill frame is
\[
T_{\mathrm{fill}} = T-k.
\]
The renderer first retrieves the full Gaussian state at $T_{\mathrm{fill}}$, which serves as the causal base scene. For Gaussians in the forecasted subset, the predicted X/S/R attributes are injected into their causal-last geometry. All uncovered Gaussians remain at their $T_{\mathrm{fill}}$ states, as illustrated in Figure~\ref{fig:pipeline}C. Opacity and SH features are also propagated from the causal-last state because the formal task forecasts geometry only. Our preliminary appearance-forecasting diagnostics show that elevating these attributes to formal prediction targets was unsatisfactory. This construction preserves a common causal base scene while allowing only predicted geometry to vary across methods.

Under this protocol, the predicted Gaussian state for frame $T$ can be summarized as
\[
\hat{\mathcal{G}}^{T}
=
(
\hat{x}^{T},
\hat{s}_{\mathrm{raw}}^{T},
\hat{r}_{\mathrm{raw}}^{T},
\alpha^{T_{\mathrm{fill}}},
c^{T_{\mathrm{fill}}}
),
\]
where $\hat{x}^{T}$, $\hat{s}_{\mathrm{raw}}^{T}$, and $\hat{r}_{\mathrm{raw}}^{T}$ are provided by the forecasting method for the forecastable trajectory subset, while $\alpha^{T_{\mathrm{fill}}}$ and $c^{T_{\mathrm{fill}}}$ are inherited from the causal-last frame. No target-frame Gaussian attributes, including $x^{T}$, $s_{\mathrm{raw}}^{T}$, $r_{\mathrm{raw}}^{T}$,
$\alpha^{T}$, or $c^{T}$, are used to construct the predicted scene. 

The target-frame camera parameters are used only to render the predicted Gaussian state from the target viewpoint, while the target image and valid mask are used only for image-space evaluation. They are not used as model inputs or as Gaussian attribute sources. Therefore, the protocol is fully causal with respect to Gaussian state and attribute access, while still enabling render-space evaluation against the target observation.

\subsection{Neural Learners and Dynamics Baselines}

All neural temporal learners regress future X/S/R residuals using an attribute-wise weighted mean-squared objective over all samples, horizon steps, and target dimensions:
\[
\mathcal{L}
=
\frac{
w_x \mathrm{SSE}_{x}
+
w_s \mathrm{SSE}_{s}
+
w_r \mathrm{SSE}_{r}
}{
w_x N_x + w_s N_s + w_r N_r
},
\]
where $\mathrm{SSE}_{x}$, $\mathrm{SSE}_{s}$, and $\mathrm{SSE}_{r}$ denote the summed squared errors for center displacement, raw scale residual, and raw rotation residual, respectively. We set $w_x=1.0$, $w_s=0.25$, and $w_r=0.25$, without horizon-specific weighting.

We instantiate $f_{\theta}$ with four neural learner families. GRU and LSTM provide recurrent baselines with gated state updates; the Transformer encoder provides an attention-based baseline; and TKAN provides a KAN-based time-series learner alternative. GRU and LSTM use three layers with hidden dimension 256, TKAN uses three TKAN layers with 256 units, and the Transformer uses hidden dimension 256, three encoder layers, eight attention heads, and feed-forward dimension 1024. All neural learners directly predict all five future steps in a single forward pass and use the same X/S/R target and full-causal-last renderer.

We also evaluate five non-learning dynamics baselines in the same X/S/R residual space: persistence, constant velocity, linear fitting, constant acceleration, and Kalman-CV. These baselines serve as controlled residual-space references rather than biomechanical tissue models; their definitions are provided in the experimental setup.

\begin{figure}[t]
\centering
\includegraphics[width=\textwidth]{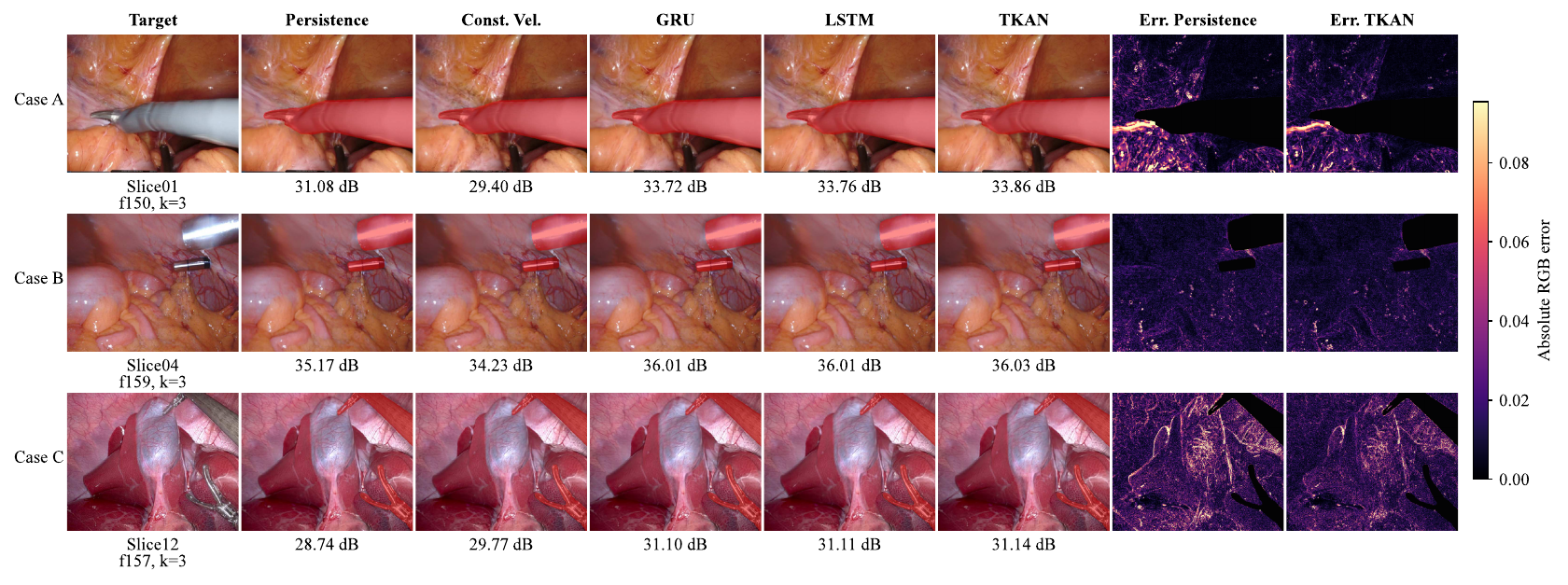}
\caption{Qualitative future-frame rendering comparison under the full-causal-last protocol at an intermediate horizon ($k=3$). The layout compares the target, persistence, constant velocity, GRU, LSTM, and TKAN forecasts, together with selected error maps. Error maps show absolute RGB differences to the target under the same valid mask; darker colors indicate smaller errors. Red regions denote surgical instrument masks, which are excluded from Deform3DGS reconstruction and rendering and are shown only for visualization.}
\label{fig:qualitative}
\end{figure}

\section{Experiments and Analysis}

\subsection{Experimental Setup and Baselines}

We evaluate SurgGMF on 12 surgical video slices from EndoNeRF \citep{wangNeuralRenderingStereo2022} and StereoMIS \citep{hayozLearningHowRobustly2023}, including one cutting sequence, one pulling sequence, and ten StereoMIS temporal segments. Most slices contain approximately 150--200 frames at 30--40 FPS, corresponding to roughly five seconds, whereas the pulling slice contains 63 frames. For each slice, a trained Deform3DGS teacher exports per-frame Gaussian states, which define the renderable state space for X/S/R forecasting. We construct temporally aligned Gaussian trajectories with history length $H=10$ and prediction horizon $P=5$, corresponding to approximately 0.25--0.33 seconds of history and 0.125--0.167 seconds of future prediction. Each slice is chronologically partitioned into
training, validation, and test subsets using a 70\%/15\%/15\% split. We retain Gaussian trajectory windows with complete and finite X/S/R histories and targets. Across all 12 video slices, more than 99.9\% of the candidate trajectory windows pass these validity checks. As summarized in Table~\ref{tab:datasets}, boundary filtering yields 1,165 legal frame-horizon pairs per forecasting method. Across nine methods, the evaluation archive contains 108 dataset--method combinations and 10,485 rendered predictions. Training was conducted on an NVIDIA GeForce RTX 4090 GPU. Model-only inference latency was measured in FP32 on the same CUDA device using batches of 10,000 Gaussian trajectories, following 20 warm-up iterations and 100 timed repetitions.

\begin{table}[t]
\centering
\small
\begin{tabular}{llcc}
\toprule
Slice & Source & Pairs / method & Rendered instances \\
\midrule
01 & EndoNeRF cutting & 75 & 675 \\
02 & EndoNeRF pulling & 25 & 225 \\
03 & StereoMIS P1A & 105 & 945 \\
04 & StereoMIS P1B & 115 & 1,035 \\
05 & StereoMIS P2-0 & 90 & 810\\
06 & StereoMIS P2-2 & 90 & 810 \\
07 & StereoMIS P2-5 & 90 & 810 \\
08 & StereoMIS P2-6A & 95 & 855 \\
09 & StereoMIS P2-6B & 110 & 990 \\
10 & StereoMIS P2-7 & 140 & 1,260 \\
11 & StereoMIS P2-8 & 115 & 1,035 \\
12 & StereoMIS P3 & 115 & 1,035 \\
\midrule
Total & -- & 1,165 & 10,485 \\
\bottomrule
\end{tabular}
\caption{Evaluation slices and valid render-space pairs. Pair counts denote valid frame--horizon pairs per forecasting method after boundary filtering; rendered instances multiply each count by the nine evaluated methods.}
\label{tab:datasets}
\end{table}

Render-space quality is evaluated with PSNR, SSIM, LPIPS, and MAE after full-causal-last rendering. PSNR, SSIM, and LPIPS follow the masked-current protocol, whereas MAE follows the valid-strict rendering protocol. These metrics assess whether predicted Gaussian motion improves future-frame rendering rather than only reducing Gaussian-parameter regression error.

All forecasting methods use the same causal history, predicted-frame export, renderer, masks, and metrics. Persistence propagates the last observed residual state, $\hat{m}^{t+k}_{i}=m^t_i$. Constant velocity extrapolates from the last two history states. Linear fit uses the full $H=10$ history window to fit a least-squares linear trend per X/S/R channel. Constant acceleration estimates second-order residual dynamics from the final history states. Kalman-CV applies a lightweight constant-velocity Kalman-style filter to the vectorized X/S/R residual state. These baselines test whether learned models add predictive value beyond hand-crafted residual extrapolation.

\subsection{Main Benchmark: Classical vs Neural Forecasting}

Table~\ref{tab:main_results} compares neural temporal learners with non-learning dynamics baselines over the 12 surgical video slices. Among the classical methods, constant velocity achieves the best PSNR, SSIM, and MAE, improving over persistence by 1.211 dB in PSNR and reducing MAE from 0.0230 to 0.0203. Persistence nevertheless obtains the best classical LPIPS, indicating that improved pixel-level alignment from linear motion extrapolation does not necessarily translate into better perceptual similarity. Constant acceleration provides only limited gains over persistence in PSNR and SSIM while degrading LPIPS and MAE, whereas linear fitting and Kalman-CV underperform persistence across all four metrics.

\begin{table}[t]
\centering
\small
\begin{tabular}{llcccc}
\toprule
Method & PSNR $\uparrow$ & SSIM $\uparrow$ & LPIPS $\downarrow$ & MAE $\downarrow$ \\
\midrule
Persistence & 29.480 & 0.8133 & 0.1818 & 0.0230 \\
Const. Vel. & 30.691 & 0.8343 & 0.1888 & 0.0203 \\
Linear fit & 28.717 & 0.7892 & 0.2004 & 0.0255 \\
Const. Accel. & 29.918 & 0.8145 & 0.2185 & 0.0234 \\
Kalman-CV & 26.638 & 0.7259 & 0.2820 & 0.0331 \\
\midrule
GRU & 31.801 & 0.8611 & 0.1768 & 0.0175 \\
LSTM & 31.810 & 0.8614 & 0.1768 & 0.0175 \\
Transformer & 31.674 & 0.8585 & 0.1778 & 0.0178 \\
TKAN & \textbf{31.895} & \textbf{0.8633} & \textbf{0.1764} & \textbf{0.0172} \\
\bottomrule
\end{tabular}
\caption{Average render-space performance over 12 surgical video slices. Higher PSNR and SSIM are better; lower LPIPS and MAE are better. Const. Vel. denotes constant velocity; Const. Accel. denotes constant acceleration.}
\label{tab:main_results}
\end{table}

Each neural learner exceeds the corresponding best classical score on every evaluated metric. Relative to the metric-wise best classical scores, the neural models improve PSNR by 0.983--1.204 dB and SSIM by 0.0242--0.0290, while reducing LPIPS by 0.0040--0.0054 and MAE by 0.0025--0.0031. These consistent gains show that learned temporal forecasting captures predictive structure beyond persistence and hand-crafted residual extrapolation.

The neural learners form a comparatively tight performance group. TKAN achieves the best numerical averages across PSNR, SSIM, LPIPS, and MAE, but its margins over GRU and LSTM are modest. LSTM and GRU produce nearly identical results, while the Transformer is slightly weaker under the evaluated short-history, short-horizon setting. The benchmark therefore supports learned Gaussian motion forecasting as a model family rather than attributing the improvement to a single temporal backbone.

Because all methods share the same causal history, predicted-state construction, renderer, masks, and metrics, Table~\ref{tab:main_results} primarily isolates differences in predicted X/S/R residual quality.

\subsection{Qualitative and Horizon Analysis}

Figure~\ref{fig:qualitative} compares future-frame renderings at the intermediate horizon ($k=3$). Across the displayed cases, learned forecasts more closely match the target in regions affected by tissue deformation and tool--tissue interaction. The selected TKAN error maps also show lower valid-region RGB errors than the classical references in the displayed examples, consistent with the aggregate results in Table~\ref{tab:main_results}.

\begin{figure}[t]
\centering
\includegraphics[width=0.8\linewidth]{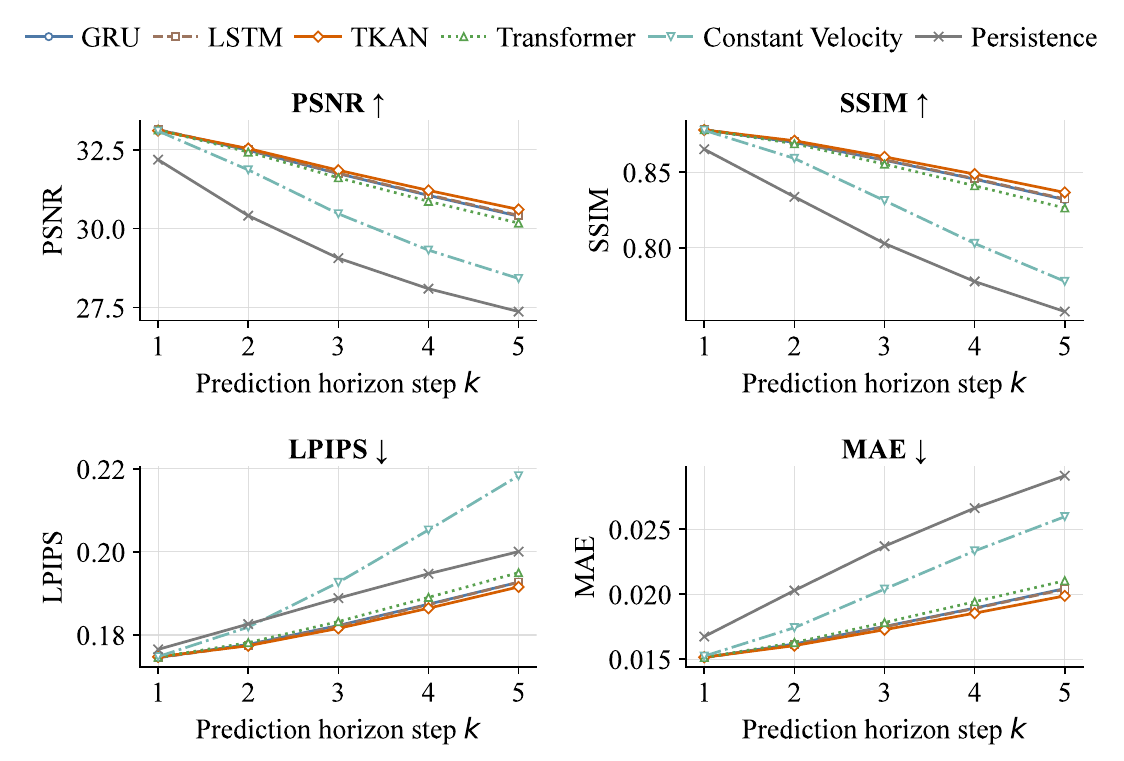}
\caption{Per-horizon render-space performance under the full-causal-last protocol. Metrics are averaged over valid frame-horizon pairs from 12 surgical video slices. Learned forecasting methods remain stronger than the classical dynamics references across the evaluated prediction horizons.}
\label{fig:per_horizon}
\end{figure}

Figure~\ref{fig:per_horizon} reports render-space performance over the five prediction steps. Performance generally decreases as the horizon increases because future states become progressively farther from the observed history. Nevertheless, the learned models retain an advantage over the classical dynamics references across the evaluated horizons, indicating that their gains are not confined to the nearest prediction step. The predicted X/S/R dynamics therefore remain useful throughout the evaluated short-term forecasting range.

Because every horizon is evaluated without target-frame Gaussian attribute access, the observed degradation reflects increasing forecasting difficulty rather than changing future-state availability. Appearance attributes and uncovered primitives remain at their corresponding causal-last states, preventing horizon-dependent quality from being inflated by target-state filling.

\subsection{Gaussian Motion Target Ablation}

We investigate how the choice of predicted Gaussian motion attributes affects future-frame rendering. For both LSTM and TKAN, we independently train four forecasting variants with different output targets: position only (X), position and scale (X+S), position and rotation (X+R), and the complete position--scale--rotation target (X+S+R). All variants use the same historical input, training data, forecasting horizon, causal rendering protocol, masks, and evaluation metrics; only the supervised output attributes differ. During inference, attributes not included in a model's forecasting target are retained from the causal-last Gaussian state.

\begin{table}[t]
\centering
\small
\begin{tabular}{llcccc}
\toprule
Model & Target & PSNR $\uparrow$ & SSIM $\uparrow$ & LPIPS $\downarrow$ & MAE $\downarrow$ \\
\midrule
LSTM & X & 30.570 & 0.8401 & 0.1798 & 0.02039 \\
LSTM & X+S & 30.900 & 0.8438 & 0.1791 & 0.01952 \\
LSTM & X+R & 31.542 & 0.8589 & 0.1774 & 0.01809 \\
LSTM & X+S+R & 31.810 & 0.8614 & 0.1768 & 0.01746 \\
\midrule
TKAN & X & 30.570 & 0.8404 & 0.1798 & 0.02037 \\
TKAN & X+S & 30.917 & 0.8442 & 0.1791 & 0.01946 \\
TKAN & X+R & 31.618 & 0.8607 & 0.1771 & 0.01787 \\
TKAN & X+S+R & \textbf{31.895} & \textbf{0.8633} & \textbf{0.1764} & \textbf{0.01723} \\
\bottomrule
\end{tabular}
\caption{Forecasting-target ablation with independently trained variants. X predicts position only; X+S adds raw scale residuals; X+R adds raw rotation residuals; X+S+R predicts all three components.}
\label{tab:attribute_ablation}
\end{table}

Table~\ref{tab:attribute_ablation} shows a consistent forecasting-target trend across both temporal backbones. Position-only forecasting alone provides a meaningful baseline, but does not fully capture the future evolution of the anisotropic Gaussian representation. Adding scale to the position target improves PSNR by 0.330 dB for LSTM and 0.347 dB for TKAN. Adding rotation produces substantially larger gains of 0.972 dB and 1.048 dB, respectively, with corresponding improvements in SSIM, LPIPS, and MAE. Under the evaluated surgical sequences, rotation is therefore more informative than scale when added conditionally to position forecasting.

The joint X+S+R target performs best for both models. Compared to X-only forecasting, it improves PSNR by 1.240 dB for LSTM and 1.325 dB for TKAN, while simultaneously yielding the highest SSIM alongside the lowest LPIPS and MAE. Its advantage over X+R shows that scale provides complementary information, despite having a smaller isolated gain over X than rotation does. The identical ordering for LSTM and TKAN indicates that the benefit of joint X/S/R forecasting is consistent across different temporal backbones rather than being specific to a single architecture.

\subsection{Accuracy--Efficiency Trade-off}

In addition to forecasting accuracy, we evaluate the computational cost of the neural temporal learners at two levels: model-only forecasting latency and minimal forecast--render latency. The first isolates temporal-learner inference, whereas the latter additionally includes history processing, predicted-state construction, and CUDA rasterization. Both are module-level measurements.

\begin{figure}[!htbp]
\centering
\includegraphics[width=0.6\linewidth]{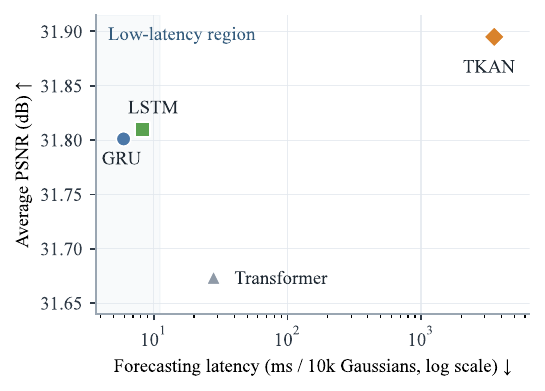}
\caption{Accuracy--efficiency trade-off for neural Gaussian motion forecasting. Average render-space PSNR is compared with model-only latency for direct five-step X/S/R prediction over 10,000 Gaussian trajectories. Latency depends on both the model and its evaluated software backend.}
\label{fig:latency_tradeoff}
\end{figure}

As shown in Figure~\ref{fig:latency_tradeoff}, GRU is the fastest model at 5.94 ms, followed by LSTM at 8.25 ms and the Transformer at 27.99 ms. TKAN achieves the highest average accuracy but requires 3,530 ms under the current Keras/JAX implementation. Its PSNR advantage over LSTM and GRU is only 0.085 and 0.094 dB, respectively. Thus, TKAN represents the accuracy-oriented operating point, whereas GRU and LSTM provide substantially more favorable module-level accuracy--latency trade-offs.

The forecast--render results confirm the efficiency advantage of recurrent learners. As shown in Table~\ref{tab:forecast_render_latency}, with AMP, GRU reaches 55.1 FPS and LSTM 41.6 FPS, compared with 14.0 FPS for the Transformer. GRU provides the lowest latency, while LSTM retains nearly identical forecasting accuracy at a moderate additional cost. Because TKAN is evaluated with Keras/JAX and the remaining learners with PyTorch, these measurements characterize the current model--implementation combinations rather than architecture-intrinsic efficiency. The reported FPS values also exclude several upstream and system-level operations and should not be interpreted as end-to-end robotic-system real-time performance.

\begin{table}[!htbp]
\centering
\small
\begin{tabular}{llcc}
\toprule
Model & Precision & Latency (ms) $\downarrow$ & FPS $\uparrow$ \\
\midrule
GRU         & FP32 & 35.08  & 28.5 \\
GRU         & AMP  & \textbf{18.14} & \textbf{55.1} \\
LSTM        & FP32 & 46.67  & 21.4 \\
LSTM        & AMP  & 24.05  & 41.6 \\
Transformer & FP32 & 151.99 & 6.6 \\
Transformer & AMP  & 71.66  & 14.0 \\
\bottomrule
\end{tabular}%
\caption{Minimal forecast--render latency for the PyTorch backbones. FPS is derived from the corresponding module latency.}
\label{tab:forecast_render_latency}
\end{table}

\FloatBarrier
\section{Conclusion}

In this paper, we present SurgGMF, a fully causal Gaussian motion forecasting framework for anticipatory surgical scene rendering. Under the full-causal-last protocol, learned temporal models outperform classical dynamics baselines, while independently trained forecasting-target ablations support joint X/S/R forecasting. Latency analysis further reveals a practical accuracy--efficiency trade-off: TKAN achieves the highest average accuracy under the current implementation, whereas GRU and LSTM provide more favorable module-level accuracy--latency trade-offs. The current study focuses on short-horizon geometry forecasting and is evaluated at the module level; future work will extend it toward longer-horizon forecasting, joint geometry--appearance prediction, and end-to-end online deployment.

\section*{Conflicts of Interest}

The authors declare that they have no conflicts of interest.

\bibliographystyle{plainnat}
\bibliography{surggmf_refs}

\end{document}